\pdfoutput=1
\documentclass[11pt, dvipsnames]{article}
\usepackage{ACL2023}
\usepackage{amsmath}
\usepackage{graphicx}

\usepackage{times}
\usepackage{latexsym}
\usepackage{amssymb}
\usepackage{hyperref}
\usepackage{booktabs}
\usepackage[T1]{fontenc}
\usepackage[utf8]{inputenc}
\usepackage{microtype}
\usepackage{inconsolata}
\title{CMLFormer: A Dual Decoder Transformer with Switching Point Learning for Code-Mixed Language Modeling}

\author{
Aditeya Baral$^{\dagger}$ \quad Allen George Ajith$^{\dagger}$ \quad 
Roshan Nayak$^{\S}$ \quad Mrityunjay Abhijeet Bhanja$^{\S}$ \\
$^{\dagger}$Courant Institute of Mathematical Sciences, $^{\S}$Tandon School of Engineering \\
New York University \\
\texttt{\{aditeyabaral, allen.ajith, rn2588, mb9348\}@nyu.edu}
}

\begin{document}
\maketitle
\begin{abstract} Code-mixed languages, characterized by frequent within-sentence language transitions, present structural challenges that standard language models fail to address. In this work, we propose \textbf{CMLFormer}, an enhanced multi-layer dual-decoder Transformer with a shared encoder and synchronized decoder cross-attention, designed to model the linguistic and semantic dynamics of code-mixed text. CMLFormer is pre-trained on an augmented Hinglish corpus with switching point and translation annotations with multiple new objectives specifically aimed at capturing switching behavior, cross-lingual structure, and code-mixing complexity. Our experiments show that CMLFormer improves F1 score, precision, and accuracy over other approaches on the HASOC-2021 benchmark under selected pre-training setups. Attention analyses further show that it can identify and attend to switching points, validating its sensitivity to code-mixed structure. These results demonstrate the effectiveness of CMLFormer’s architecture and multi-task pre-training strategy for modeling code-mixed languages. \end{abstract}

\section{Introduction}

Globalization and cultural adaptation have fostered the growth of multilingualism in the modern world. This has led to a widespread phenomenon of \textit{code-mixing} (CM), where multiple languages are juxtaposed within a single utterance, with the transition points being known as \textit{switching points} (SP). In multilingual countries like India, the fusion of Hindi and English (popularly termed \textit{Hinglish}) has become increasingly prevalent in conversational usage both online and offline. In code-mixed languages, the \textit{base language} provides the grammatical structure, while the \textit{mixing language} contributes loan words or phrases. For example, for the CM sentence  
$\textit{Mujhe}_\text{HI}$ $\textit{kal}_\text{HI}$ $\textit{ek}_\text{HI}$ $\textit{important}_\text{EN}$ $\textit{meeting}_\text{EN}$ $\textit{attend}_\text{EN}$ $\textit{karni}_\text{HI}$ $\textit{hai}_\text{HI}$,  
Hindi serves as the base language, while the words \textit{important}, \textit{meeting}, and \textit{attend} from English are mixed in.


As Large Language Models (LLMs) become increasingly integrated into our daily lives, recent years have seen the development of multiple LLMs for a range of diverse languages, including high-resource ones such as English \cite{geminiteam2024geminifamilyhighlycapable} and French \cite{faysse2024croissantllmtrulybilingualfrenchenglish}. Despite the steady increase in CM speakers, most LLMs still fail to serve CM needs \cite{srivastava-2025-dweshvaani, venkatesh-etal-2024-bits}. This is evidenced by the subpar performance on real-world tasks requiring encoder-only models, such as classification, semantic search and information retrieval. Their limitations can be traced back to their inadequate and poor neural representations of these languages \cite{mazumder2024revealingimpactsyntheticnative, jagdale2024importancecodemixedembeddingshate}. The unique linguistic structure of CM languages makes it a challenge to represent them effectively, limiting their efficacy in multilingual contexts. 

\begin{figure}
    \centering
    \includegraphics[width=1\linewidth]{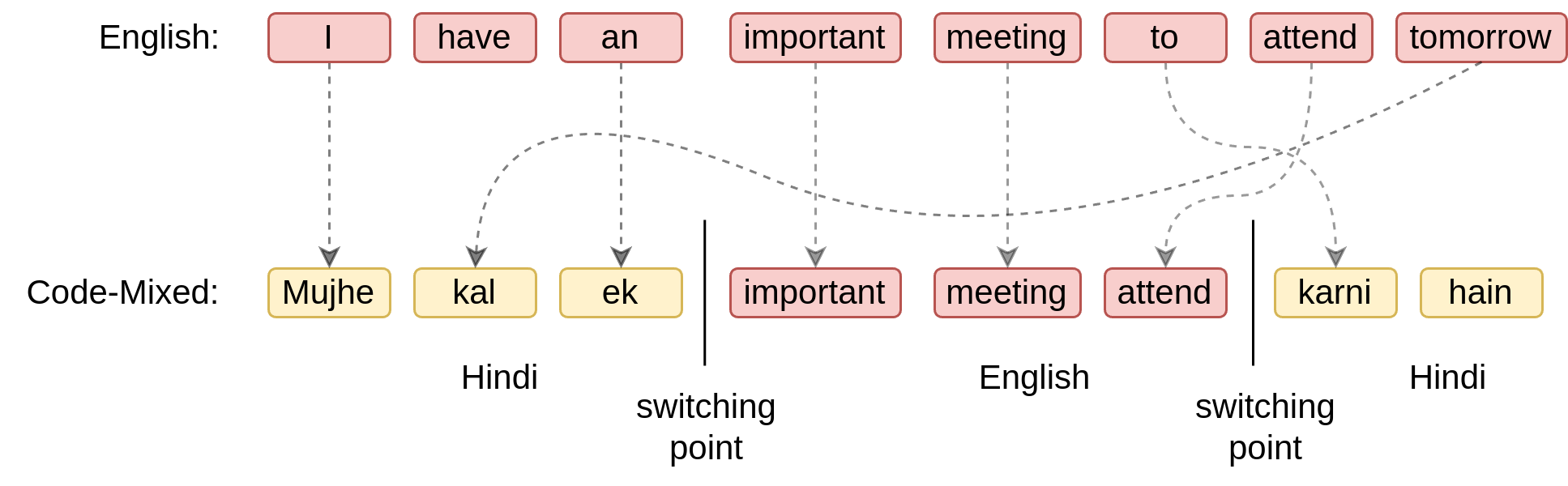}
    \caption{\textbf{Switching Points in Code-Mixed Text alter structural dynamics.} Code-mixing introduces \textit{switching points} where the language transitions mid-sentence, significantly affecting token order and grammatical structure. In the example shown, words like \textit{tomorrow} (originally after \textit{attend} in English) now appear earlier in the Hindi-English mixed variant, changing the sentence's grammatical flow, which multilingual models struggle to capture.}
    \label{fig:cmsp}
\end{figure}

Addressing these limitations requires a richer modeling of CM structure, particularly the role of switching points in transitional contexts. Since prior studies suggest that current pre-training approaches are insufficient to model CM languages, we propose \textbf{CMLFormer}, a novel dual-decoder Transformer architecture with switching point learning, specifically for code-mixed language modeling. Additionally, we introduce new supervised and unsupervised pre-training objectives such as \textit{Switching Point Prediction} and \textit{Bilingual Language Translation Modeling} to learn richer representations of CM languages and encourage cross-lingual learning and alignment.

The main contributions of this work are summarized as follows,
\begin{itemize}
\setlength{\itemsep}{-1pt}
    \item An enhanced multi-layer Transformer with a shared encoder and synchronized coupling between of two decoders for CM language modeling.
    \item Multiple new pre-training objectives to capture syntactic and semantic dynamics of language transitions in CM text.
    \item Augmentation of the L3Cube-HingCorpus dataset with labels for switching points and translations of CM sentences.
    \item Extrinsic and intrinsic evaluation of our work against baseline models on recognised CM classification benchmarks.
\end{itemize}

\section{Related Work}

Previous approaches have used multilingual models to bridge the gap between the languages in a CM language. However, studies have shown that CM languages being \textit{inherently} multilingual cannot be effectively modeled by multilingual models since they are not natural code-mixers \cite{zhang2023multilinguallargelanguagemodels} of monolingual languages. The lack of formal grammar, frequent occurrence of switching points, spelling variants and contextual nuances are some of the problems posed by CM languages. Additionally, models pre-trained on CM data have surprisingly shown inadequate improvements over multilingual models despite increasing the number of parameters \cite{Patil_2023, santy-etal-2021-bertologicomix}. Since scaling up both data and parameters has only yielded marginal improvements, recent studies have moved away from the former and established the necessity of specialised pre-training techniques for CM languages. 

\cite{li2022languageagnosticcodemixingdata} introduce a language-agnostic approach that shows that masking the loan words from a CM sequence allows multilingual models to generalize better to downstream tasks. The effectiveness of cross-lingual representations using approaches like Translation Language Modeling \cite{lample2019crosslinguallanguagemodelpretraining} has also been extensively studied in MuRIL \cite{khanuja2021murilmultilingualrepresentationsindian} and IndicBERT \cite{doddapaneni2023leavingindiclanguagebehind}, both of which demonstrate significant improvements over multilingual BERT \cite{devlin2019bertpretrainingdeepbidirectional} on Indian languages. However, these approaches still failed to sufficiently capture the linguistic nuances of CM text. Moving towards architectural modifications, \cite{sengupta-etal-2021-hit} introduce a Hierarchical Transformer with a new outer-product attention mechanism to effectively capture the syntactic and structural characteristics of CM sentences. \cite{ali2021pestoswitchingpointbased} was one of the first to introduce a new positional encoding approach that assigned weights to words near switching points, enabling the model to capture language transitions in a CM sentence better. Building upon this, \cite{ali2023conflatorincorporatingswitchingpoint} introduce an improved switching point based
rotary matrix by combining it with rotatory positional encodings \cite{su2023roformerenhancedtransformerrotary}, which rotate at switching points to denote language transitions.

\section{CMLFormer}

This section introduces our proposed model, the CMLFormer, and includes details about the architecture, pre-training tasks, data, and other details.

\subsection{Problem Definition}


We formally represent a sequence of $N_C$ tokens in the CM language as $C = \{c_1, c_2, \dots, c_{N_C}\}$, a sequence of translated $N_B$ tokens in the base language as $B = \{b_1, b_2, \dots, b_{N_B}\}$, and a sequence of translated $N_M$ tokens in the mixing language as $M = \{m_1, m_2, \dots, m_{N_M}\}$. We also define the set of language labels with $L = \{l_1, l_2, \dots, l_{N_C}\}$ and language transitions with a bit vector $T = \{t_1, t_2, \dots, t_{N_C}\}$, where $t_i$ is a 0-1 bit that indicates a language transition \textit{before} token $c_i$, that is, $t_i = 1$ if $l_i \neq l_{i-1}$, and $0$ otherwise. The vector always starts with a $0$ at the beginning since there is no transition on the first token.

For example, for an input CM sentence $C$: $\textit{college}_\text{EN}$  $\textit{mein}_\text{HI}$ $\textit{aaj}_\text{HI}$ $\textit{exam}_\text{EN}$  $\textit{hai}_\text{HI}$, we denote the language labels $L = \{\text{EN}, \text{HI}, \text{HI}, \text{EN}, \text{HI}\}$ and transitions $T = \{0, 1, 0, 1, 1\}$. The input to CMLFormer's encoder is the sequence $C$, with $B$ and $M$ being passed to the base and mixing language decoder, respectively.

\subsection{Architecture}

\begin{figure}[ht]
    \centering
    \includegraphics[width=1\columnwidth]{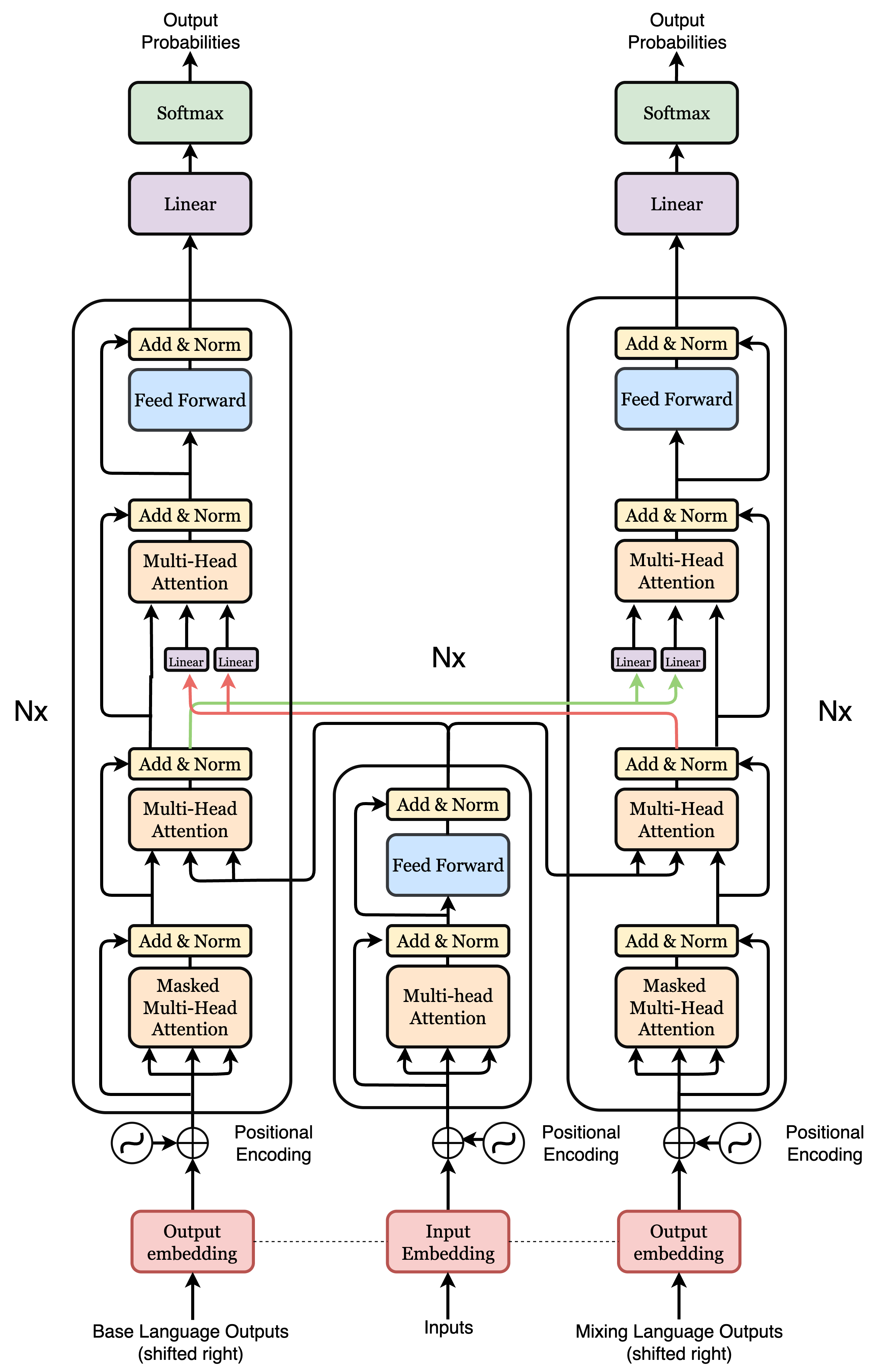}
    \caption{\textbf{The architecture of our proposed approach CMLFormer}:
    The outputs from each encoder-decoder attention sub-layer (arrows shown in \textcolor{YellowGreen}{green} and \textcolor{Bittersweet}{red}) are passed as input to the decoder-decoder cross-attention sub-layer. The decoders exhibit full synchronous coupling since each requires the hidden states from the other to compute its own hidden states. After pre-training, the encoder is extracted and fine-tuned on downstream tasks.}
    \label{fig:cmlformer-architecture}
\end{figure}

We build the backbone of the CMLFormer (Fig. \ref{fig:cmlformer-architecture}) using an enhanced multi-layer Transformer \cite{vaswani2017attention} with both the encoder and decoder layers. We propose two significant modifications to the vanilla Transformer architecture. First, we explore a novel multi-target pre-training setup where two fully synchronous decoders sharing the same encoder are coupled through an attention layer. Considering the usually significant syntactical and semantic differences between the base and mixing language, this allows the model to decouple the two languages and handle their specific nuances effectively while still learning cross-lingual representations. We further explore two variations of this enhancement through our ablations (Appendix \ref{app:architecture-ablation}). 

Second, to help the architecture capture language transitions and learn richer representations of code-mixed inputs, we introduce additional pre-training objectives focused on code-mixing that teach the encoder to recognize where and how language transitions occur within a sentence. These tasks are built directly into the model through auxiliary prediction layers on top of the encoder, allowing it to learn not just from the main language modeling and translation tasks, but also from signals about language identity, switching points, and mixing patterns, enabling the encoder to learn richer representations of the structure of code-mixed text during training.

\subsubsection{Dual Decoder with Cross Attention}

We introduce a synchronous coupling of two decoders with the same shared encoder. In contrast to the vanilla Transformer architecture, we add a third attention layer to introduce inter-decoder cross-attention in each decoder block to mutually share the latent characteristics of each constituent language. Denoting the output hidden states from layer normalization after the encoder-decoder attention sub-layer of the base and mixing language decoder as $a^{b}_{l}$ and $a^{m}_{l}$ respectively, the decoder-decoder cross-attention is computed as,

\begin{equation}
\begin{split}
o_{l}^b &= {Attention}(a_l^b, a_l^m, a_l^m) \\
o_{l}^m &= {Attention}(a_l^m, a_l^b, a_l^b)
\end{split}
\end{equation}

The corresponding output attention scores $o_{l}^b$ and $o_{l}^m$ are then passed as input to the feed-forward and layer normalization sub-layers as in the vanilla Transformer. This allows each decoder to attend to and "peek" at the other decoder's hidden states while decoding the input CM sentence into their corresponding language, helping it capture inter-block interactions and driving cross-lingual learning and alignment. The two decoders are thus fully synchronized as each requires the hidden states of the other in each block to compute its own.

\subsection{Cross-Decoder Linear Projections}

To facilitate effective cross-attention between the base and mixing decoders, we introduce cross-decoder linear projections. Before applying decoder-decoder cross-attention, each decoder's hidden states are linearly projected into the latent space of the other decoder. This allows the base decoder to attend to representations expressed in the mixing decoder’s feature space, and vice versa. By adapting the representations across decoders, the projections account for structural and semantic asymmetries between the two languages, enabling more coherent cross-lingual interaction during decoding.



\subsection{Pre-Training Tasks}
\label{sec:pre-training-tasks}

CMLFormer is expected to learn rich representations of the CM language, including cross-lingual representations for the base and mixing language, and the relationship between tokens at switching points. To achieve this, we introduce multiple supervised and unsupervised pre-training tasks to learn rich contextual representations.

\subsubsection{Masked Language Modeling}

We incorporate the widely used Masked Language Modeling \cite{devlin2019bertpretrainingdeepbidirectional} task as the base pre-training objective. Masked Language Modelling (MLM) was first introduced in BERT to mitigate the problem of traditional bidirectional language modelling, allowing the model to "see" the tokens being predicted. Given a token sequence $S = \{s_1, s_2, ..., s_{N_S}\}$, a subset of tokens are randomly sampled denoted by $L_{mask}$, typically 15\% of which 80\% are replaced by $[MASK]$ token, 10\% are replaced by a random token, and 10\% remain unchanged. The objective of MLM is to predict the masked tokens by using the surrounding unmasked tokens in the input sequence as context. We pre-train on the MLM objective using all three $C$, $B$ and $M$ language sequences and average over the three cross-entropy losses $\mathcal{L}_{MLM_{c}}$, $\mathcal{L}_{MLM_{b}}$ and $ \mathcal{L}_{MLM_{m}}$.





\subsubsection{Bilingual Translated Sentence Prediction}
The multilingual setting of CM languages enables concepts to be represented by words from the base or mixing languages. It is necessary to model cross-lingual relationships between these word variants by learning rich contextual representations of the CM, base and mixing languages. We propose a new Bilingual Translated Sentence Prediction (BTSP) objective to capture these cross-lingual relationships. Given a CM token sequence $C$, we randomly sample another token sequence $S$ with 50\% positive examples (25\% of the time $S$ is $B$, 25\% of the time it is $M$) and 50\% negative examples (25\% of the time it is a randomly selected sequence $B'$ and the remaining 25\% of the time it is a randomly selected sequence $M'$ from the dataset). The objective of BTSP, formulated using binary cross-entropy as $\mathcal{L}_{BTSP}$, is to predict whether sequence $C$ is equivalent to the sampled sequence $S$.



\subsubsection{Bilingual Language Translation Modeling}

To capture the inherent multilingual characteristics of CM languages, it is crucial to not only learn rich contextual representations for the CM, base and mixing languages but also model a three-way alignment among them. Given a CM token sequence $C$, and its corresponding translations in the base and mixing language translations $B$ and $M$ respectively, we introduce a multi-target Bilingual Language Translation (BiLTM) objective. The base and mixed language decoders are tasked to simultaneously generate translations of the CM sequence into its equivalent base language and mixing language forms $L_B$ and $L_M$, respectively. The objective of BiLTM can be formulated as the average over the autoregressive modeling losses $\mathcal{L}_{BiLTM_{b}}$ and $\mathcal{L}_{BiLTM_{m}}$ for each decoder.



\subsubsection{Token Language Classification}

Code-mixing enables words to be shared from its constituent languages, also known as \textit{lexical borrowing}. These \textit{loan words} can thus occur within the context of a base or mixing language sentence, with each having a different syntactic and lexical structure within its local language's context. CM representations must capture this phenomenon to effectively represent the mixing of two languages and thus, be able to identify the language in which a word occurs from a given local context window. 

We propose a Token Language Classification (TLC) objective $\mathcal{L}_{TLC}$ using binary cross-entropy to predict the language in which each token $w_i$ occurs from its given local context. We construct a concatenated sequence $S$ comprising $C$, $B$ and $M$ in a randomly shuffled order. This concatenation allows \textit{loan words} to concurrently occur in two out of the three segments of $S$. Consequently, the model must rely solely on local context while predicting, enabling it to learn the usage of \textit{loan words} across different languages and linguistic structures. 



\subsubsection{Switching Point Prediction}

Switching points mark the locations where a language transition occurs between consecutive tokens in a code-mixed sequence. Modeling these transitions explicitly allows the encoder to capture the structural dynamics of language alternation, which are often critical to understanding the syntactic and semantic properties of code-mixed text. 

Given a code-mixed sequence $C = \{c_1, c_2, \dots, c_{N_C}\}$ and a corresponding switching point label sequence $T = \{t_1, t_2, \dots, t_{N_C}\}$, where $t_i = 1$ if a switch occurs between $c_i$ and $c_{i-1}$ and $0$ otherwise, the task is to predict the probability $p_i$ of a switch occurring at each token position. We treat the SPP objective $\mathcal{L}_{\mathrm{SPP}}$ as a token-level binary classification task, where the loss is computed using binary cross-entropy between the predicted switching probabilities $p_i$ and the ground truth labels $t_i$.



\subsubsection{Code-Mixing Index Prediction}
The Code Mixing Index (CMI) is a commonly used metric that quantifies the degree of code-mixing in a sentence by measuring the relative proportion of tokens across its constituent languages \cite{das2014identifying}. Given a sequence of $N$ tokens, with $P$ switching points and $N_d$ tokens from the base language, we adapt the CMI formulation for our objective as, 

\begin{equation} CMI = \frac{w_n \cdot (N - N_d) + w_p \cdot P}{N} \end{equation}

where $w_n$ and $w_p$ are weighting factors. This extension balances the contribution of language distribution and the frequency of transitions in the code-mixed sentence.

To encourage the encoder to model these sentence-level code-mixing dynamics, we introduce CMI Prediction as a regression objective. The encoder is tasked with predicting a scalar $\hat{c}_C$ for each input sequence $C$, with the ground truth $c_C$ computed directly from the labels, which is then optimized using Mean Squared Error (MSE). This task complements local objectives such as Switching Point Prediction and Token Language Classification by providing a global signal for code-mixing complexity, capturing both global mixing and local switching behavior.

\subsubsection{Overall Loss Objective}

CMLFormer is jointly trained on all the above pre-training objectives to minimise the overall training loss $\mathcal{L}_{total}$. This can be formulated as,

\vspace{-10pt}
\begin{equation}
\begin{split}
\mathcal{L}_{total} = 
\alpha \mathcal{L}_{MLM} 
+ \beta \mathcal{L}_{SPP} 
+ \gamma \mathcal{L}_{BTSP} \\
+ \eta \mathcal{L}_{BiLTM} 
+ \zeta \mathcal{L}_{TLC} 
+ \delta \mathcal{L}_{CMI}
\end{split}
\end{equation}

where $\alpha$, $\beta$, $\gamma$, $\eta$, $\zeta$, and $\delta$ are scaling constants.

\subsection{Datasets}
\subsubsection{Pre-training}
CMLFormer is pre-trained on the L3Cube-HingCorpus \cite{nayak-joshi-2022-l3cube}, a large-scale Hindi-English code-mixed corpus in Roman script. It comprises 52.93M\footnote{Due to computational limitations and time constraints, it was not possible to augment and pre-train on all 52.3M sentences. Instead, we sampled 10,000 sentences for pre-training.} sentences and 1.04B tokens from Twitter and reflects real-world code-mixing patterns. To support our auxiliary pre-training objectives, we augment the dataset by generating parallel translations of code-mixed Hinglish sentences into pure Hindi and English in Roman script and annotate each code-mixed sentence with its corresponding switching points. More details about the data augmentation can be found in Appendix \ref{app:dataset-aug-prep}.

\subsubsection{Fine-tuning and Evaluation}
CMLFormer is evaluated on downstream tasks by detaching the decoders and fine-tuning just the encoder on an established Hinglish CM language classification benchmark, the \textbf{HASOC 2021} \cite{mandl2021overviewhasocsubtrack2021} dataset for hate speech detection. It comprises 7,000 code-mixed Hinglish tweets distributed across 5,740 train and 1,348 test examples. We compare its performance against HingBERT, a code-mixed $\text{BERT}_{\text{base}}$ model for Hinglish with MLM as its only pre-training objective.\footnote{We pre-trained $\text{BERT}_{\text{base}}$ on our sample of the dataset to produce a comparable HingBERT to benchmark against.} 

\section{Pre-training $\text{CMLFormer}_{\text{base}}$\footnote{For our base model, we restricted the encoder size to match $\text{BERT}_{\text{base}}$, which can be found in Table \ref{tab:model-config-base}. This ensures that the number of parameters in the encoder remains consistent with the BERT-based model we will benchmark against.}}

\subsection{Training a Tokenizer}

To support our pre-training objectives, we trained a custom WordPiece tokenizer using the combined augmented dataset, i.e, parallel sequences in code-mixed Hinglish, base Hindi, and mixing English. Rather than training separate tokenizers for each language, we use a single tokenizer with a shared vocabulary across all three. This design choice ensures consistent subword segmentation across the languages and simplifies the overall architecture. Our trained tokenizer can be found on HuggingFace\footnote{\href{https://huggingface.co/cmlformer/L3Cube-HingCorpus10K}{https://huggingface.co/cmlformer/L3Cube-HingCorpus10K}}.

\subsection{Multi-task Pre-training and Optimization}

We pre-train CMLFormer with all pre-training objectives jointly enabled, allowing the encoder to optimize across multiple linguistic and structural tasks simultaneously. During training, we monitor the loss behavior of each objective individually to ensure stable multi-task optimization. The full model configuration and pre-training hyperparameters are provided in Appendix~\ref{app:pretraining-details}. 
 
\section{Fine-tuning}
\label{sec:fine-tuning}

During fine-tuning, we detach the dual decoders and attach a task-specific classification head to the encoder. We perform full fine-tuning, i.e, both the encoder and classification head parameters are updated jointly\footnote{We repeat the same for the $\text{BERT}_{\text{base}}$ model pre-trained on our dataset}. Our fine-tuning hyperparameters are provided in Appendix \ref{app:finetuning-details}. CMLFormer's implementation with its pre-training and fine-tuning scripts can be found here\footnote{\href{https://github.com/cmlformer/cmlformer}{https://github.com/cmlformer/cmlformer}}.

\section{Results}

\begin{table*}[htp]
\centering
\resizebox{1\textwidth}{!}{%
\begin{tabular}{l | c c c c c c | c c | c c | c c | c c}
\hline
\textbf{Model} & \textbf{MLM} & \textbf{BiLTM} & \textbf{SPP} & \textbf{BTSP} & \textbf{TLC} & \textbf{CMI} & 
\textbf{Precision} & $\Delta$ & \textbf{Recall} & $\Delta$ & \textbf{Accuracy} & $\Delta$ & \textbf{F1} & $\Delta$ \\ 
\hline
$\text{BERT}_{\text{base}}$ & \checkmark & & & & & & 0.189 & -- & 0.367 & -- & 0.496 & -- & 0.249 & -- \\ 
\hline
\multicolumn{15}{c}{} \\[0.5em] 
\hline
& \checkmark & \checkmark & & & & & \textbf{0.327} & \textbf{\textcolor{green!60!black}{$\uparrow\!0.138$}} & \textbf{0.633} & \textbf{\textcolor{green!60!black}{$\uparrow\!0.273$}} & \textbf{0.504} & \textbf{\textcolor{green!60!black}{$\uparrow\!0.008$}} & \textbf{0.431} & \textbf{\textcolor{green!60!black}{$\uparrow\!0.182$}} \\ 
& \checkmark & \checkmark & \checkmark & & & & 0.223 & \textcolor{green!60!black}{$\uparrow\!0.034$} & 0.433 & \textcolor{green!60!black}{$\uparrow\!0.066$} & 0.498 & \textcolor{green!60!black}{$\uparrow\!0.002$} & 0.295 & \textcolor{green!60!black}{$\uparrow\!0.046$} \\ 
$\text{CMLFormer}_{\text{base}}$ & \checkmark & \checkmark & \checkmark & \checkmark & & & 0.086 & \textcolor{red}{$\downarrow\!0.103$} & 0.167 & \textcolor{red}{$\downarrow\!0.200$} & 0.490 & \textcolor{red}{$\downarrow\!0.006$} & 0.113 & \textcolor{red}{$\downarrow\!0.136$} \\ 
& \checkmark & \checkmark & \checkmark & \checkmark & \checkmark & & 0.120 & \textcolor{red}{$\downarrow\!0.069$} & 0.233 & \textcolor{red}{$\downarrow\!0.134$} & 0.492 & \textcolor{red}{$\downarrow\!0.004$} & 0.159 & \textcolor{red}{$\downarrow\!0.090$} \\ 
& \checkmark & \checkmark & \checkmark & \checkmark & \checkmark & \checkmark & 0.155 & \textcolor{red}{$\downarrow\!0.034$} & 0.300 & \textcolor{red}{$\downarrow\!0.067$} & 0.494 & \textcolor{red}{$\downarrow\!0.002$} & 0.204 & \textcolor{red}{$\downarrow\!0.045$} \\ 
\hline
\end{tabular}
}
\caption{\textbf{Results on HASOC 2021 across different pre-training objectives.} CMLFormer outperforms the baseline $\text{BERT}_{\text{base}}$ across all metrics when CLM and SPP pre-training strategies are applied, while we see a minor drop in performance when other objectives are further added. A \checkmark indicates the pre-training strategy applied, \textbf{bold} indicates the best performance on that metric, and \textcolor{green!60!black}{green} and \textcolor{red}{red} indicate a gain or drop in performance, respectively.}
\label{tab:hasoc-results}
\end{table*}

\subsection{Pre-training}

\subsubsection{Complexity of Pre-training Tasks and Role of Encoder}

\textbf{Pre-training Complexity}. We observe that CMLFormer converges quickly on most tasks before reaching a long stagnating tail. The lack of improvement beyond the first epoch could be attributed to the large number of pre-training objectives, increasing the complexity of the overall pre-training stage. The small size of the pre-training dataset also contributes to a lack of robust generalization across training examples.

We observe a steady decrease in the decoder-based losses $\mathcal{L}_{{BiLTM}_{b}}$ and $\mathcal{L}_{{BiLTM}_{m}}$ before stagnating while the encoder-based losses $\mathcal{L}_{{MLM}_{c}}$, $\mathcal{L}_{{MLM}_{b}}$ and $\mathcal{L}_{{MLM}_{m}}$ do not show much convergence before stagnating. This could hint towards the higher overall complexity of the encoder-based objectives compared to decoder-based ones. Additionally, the encoder is responsible for not only generalizing well to its own tasks but also learning rich representations that aid the decoder in autoregressively generating sequences. 

\textbf{Encoder Optimization}. Since the encoder is optimized for multiple training objectives, it bears a heavier representational burden and faces a greater optimization challenge than each decoder, which is trained for a single task. Consequently, the encoder's loss exhibits slower convergence during training. Due to its involvement in multiple training objectives, the base variant's encoder might not be robust enough and could benefit from a larger model with a higher number of parameters. We will explore this in our future experiments, where we restrict our encoder to match the parameters in $\text{BERT}_{\text{large}}$ while retaining the same size of the decoder.

\subsection{Fine-tuning}

Our fine-tuning results (Table \ref{tab:hasoc-results}) show the efficacy of our auxiliary CM-focused pre-training objectives on a downstream classification task. We observe that additional pre-training tasks such as \textit{Bilingual Translation Language Modeling} and \textit{Switching Point Prediction} significantly outperform the $\text{BERT}_{\text{base}}$ model pre-trained on the same dataset with an improvement of 0.18 and 0.05 on the F1 metric, respectively. This enforces our hypothesis that learning cross-lingual aligned representations of the constituent languages and understanding the structural dynamics of switching points helps CMLFormer model code-mixed inputs more effectively than other approaches. 

However, contrary to our intuition, we see that further adding other pre-training objectives like BTSP, TLC and CMI degrade the overall performance of the model. This decline may be attributed to conflicting learning signals introduced by these additional objectives during pre-training, which could make optimization harder and reduce the model's ability to generalize effectively. We hypothesize that increasing the size of the encoder to handle the representational burden from the large number of pre-training tasks, along with pre-training on a larger dataset to generalize to more code-mixed texts, could help the model perform better on downstream tasks.

\subsection{Learning Switching Point Dynamics}
\label{sec:attention-analysis}

We visualize attention scores generated by CMLFormer on different code-mixed examples to understand the model's focus on switching points and language transitions. Selected examples are shown in Figure \ref{fig:att_main}, with additional visualizations provided in Appendix~\ref{app:attention-appendix}. We plot the scaled average attention score per token from CMLFormer's encoder's first self-attention layer and compare the corresponding scores from BERT. Choosing the first attention head and the early layers allows us to investigate the sensitivity of each model to local linguistic patterns, like language transitions at switching points.

Our results show that CMLFormer far exceeds $\text{BERT}_{\text{base}}$'s capability at identifying switching points in code-mixed inputs. Our switching point focused pre-training objectives enable it to consistently attend to language transitions, giving it significantly higher weights compared to other tokens. We also observe that CMLFormer is agnostic to the \textit{number} as well as the \textit{nature} of switching points (base-to-mix or mix-to-base), being confidently able to scale up attention around these tokens while BERT fails to identify any kind of switching point in the input. 

These results reinforce that by explicitly modeling switching behavior, our model can learn to internalize patterns that traditional models like BERT overlook. The consistency of attention alignment around transition points, regardless of directionality or position, highlights both the effectiveness of the multi-task objectives and the representational advantages of CMLFormer's design.

\begin{figure*}[h]
\centering

\begin{minipage}{0.47\textwidth}
    \centering
    \includegraphics[width=\linewidth]{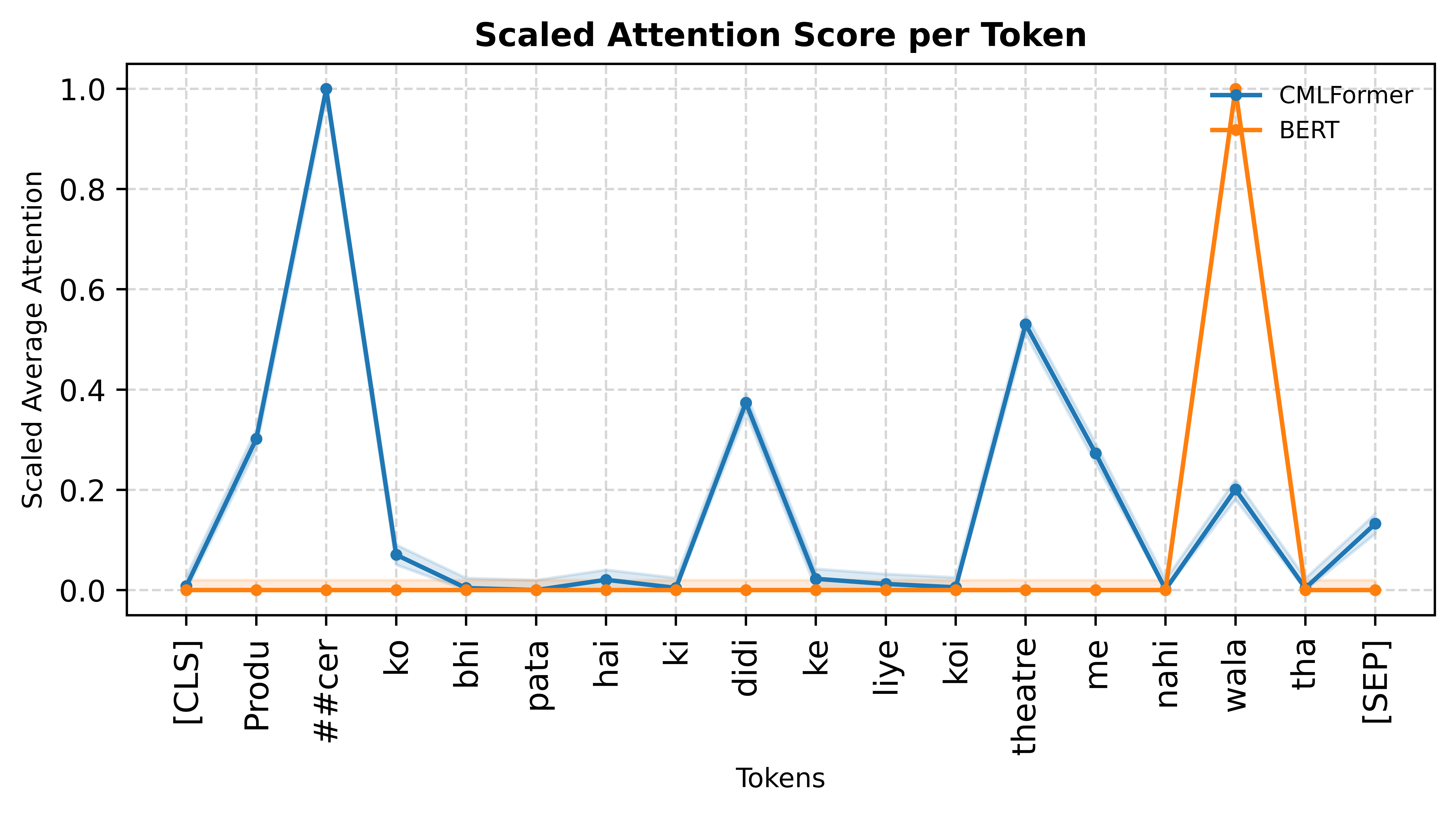}
\end{minipage}
\hfill
\begin{minipage}{0.47\textwidth}
    \centering
    \includegraphics[width=\linewidth]{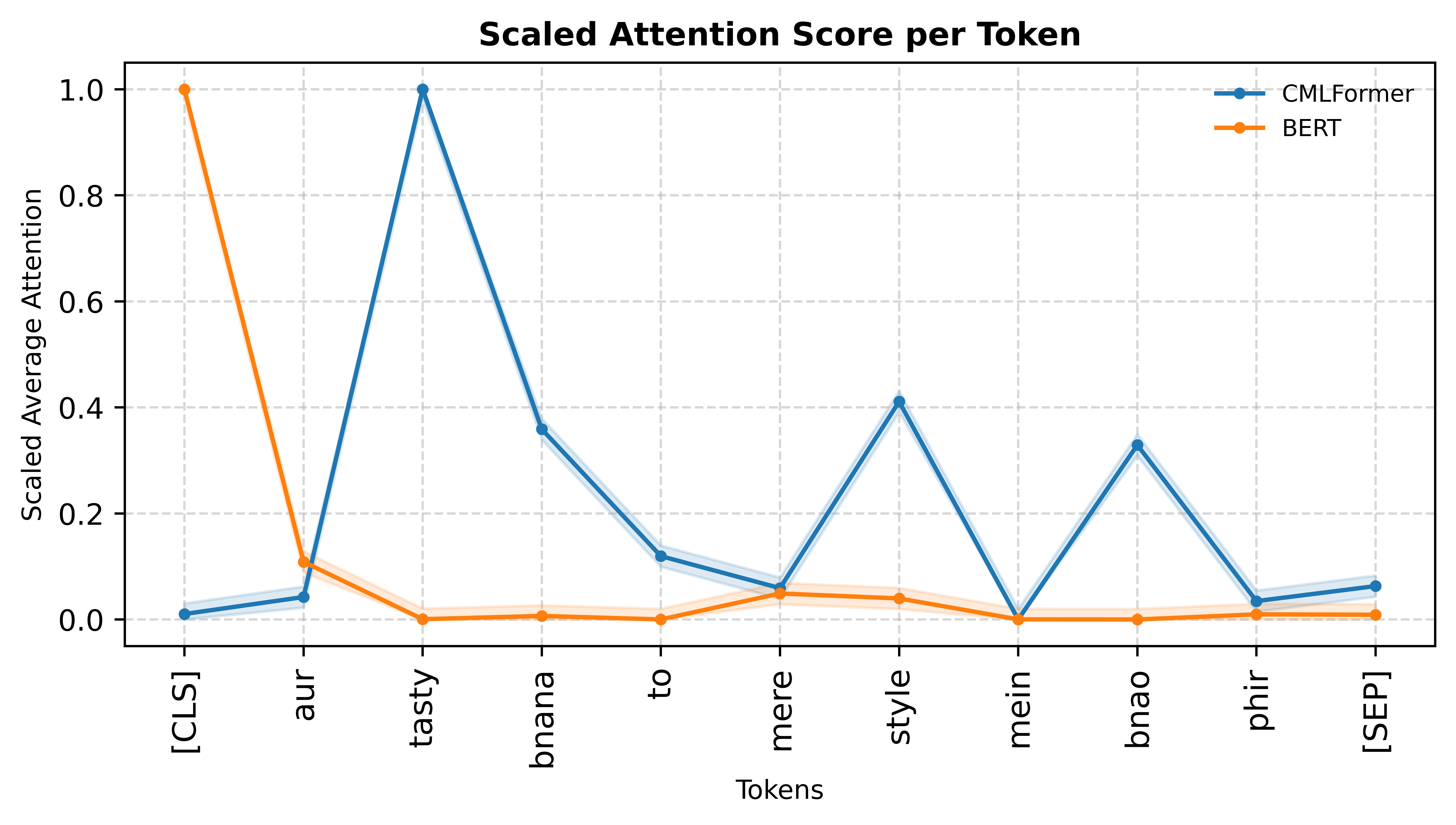}
\end{minipage}

\vspace{1em}

\begin{minipage}{0.47\textwidth}
    \centering
    \includegraphics[width=\linewidth]{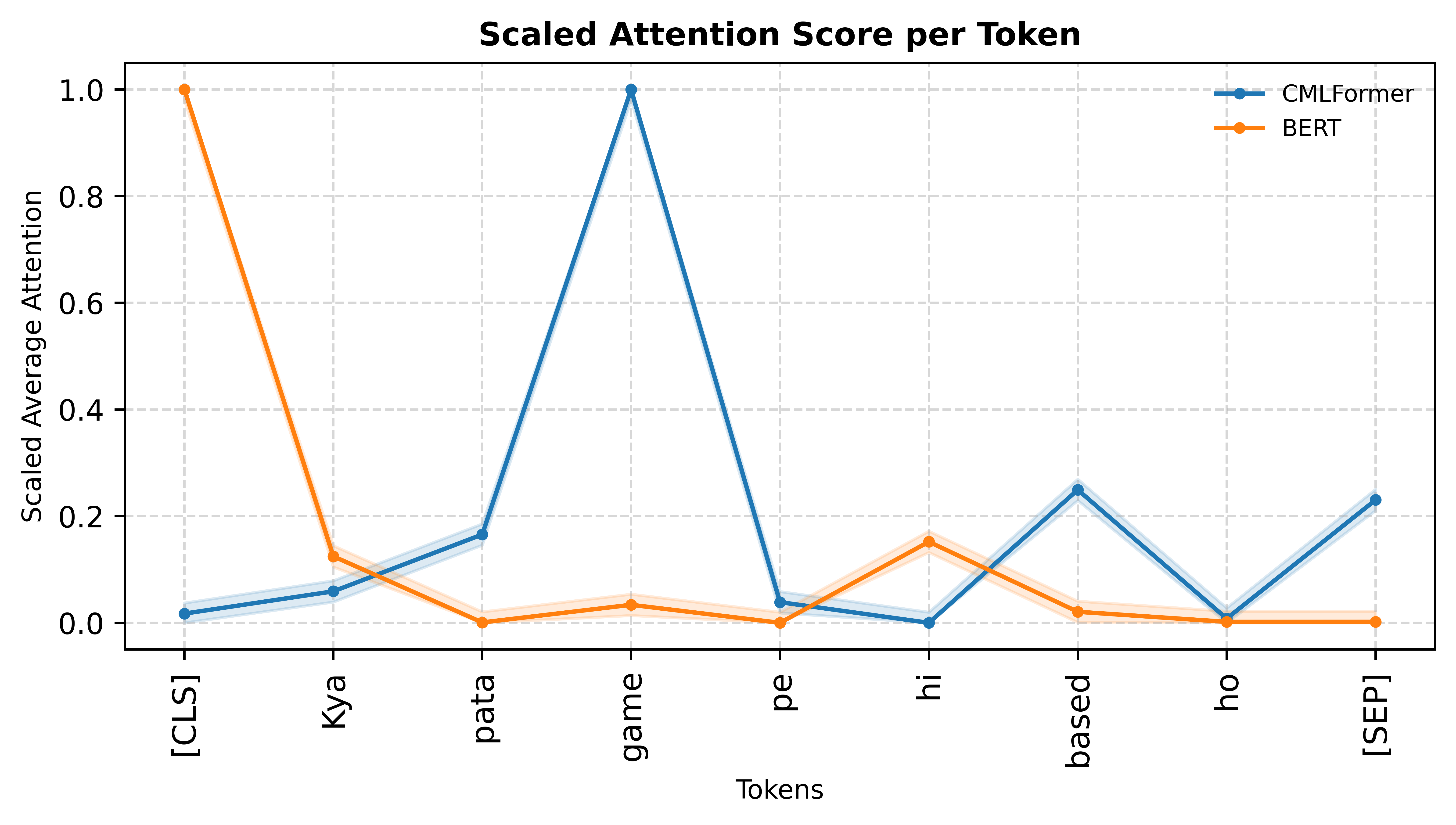}
\end{minipage}
\hfill
\begin{minipage}{0.47\textwidth}
    \centering
    \includegraphics[width=\linewidth]{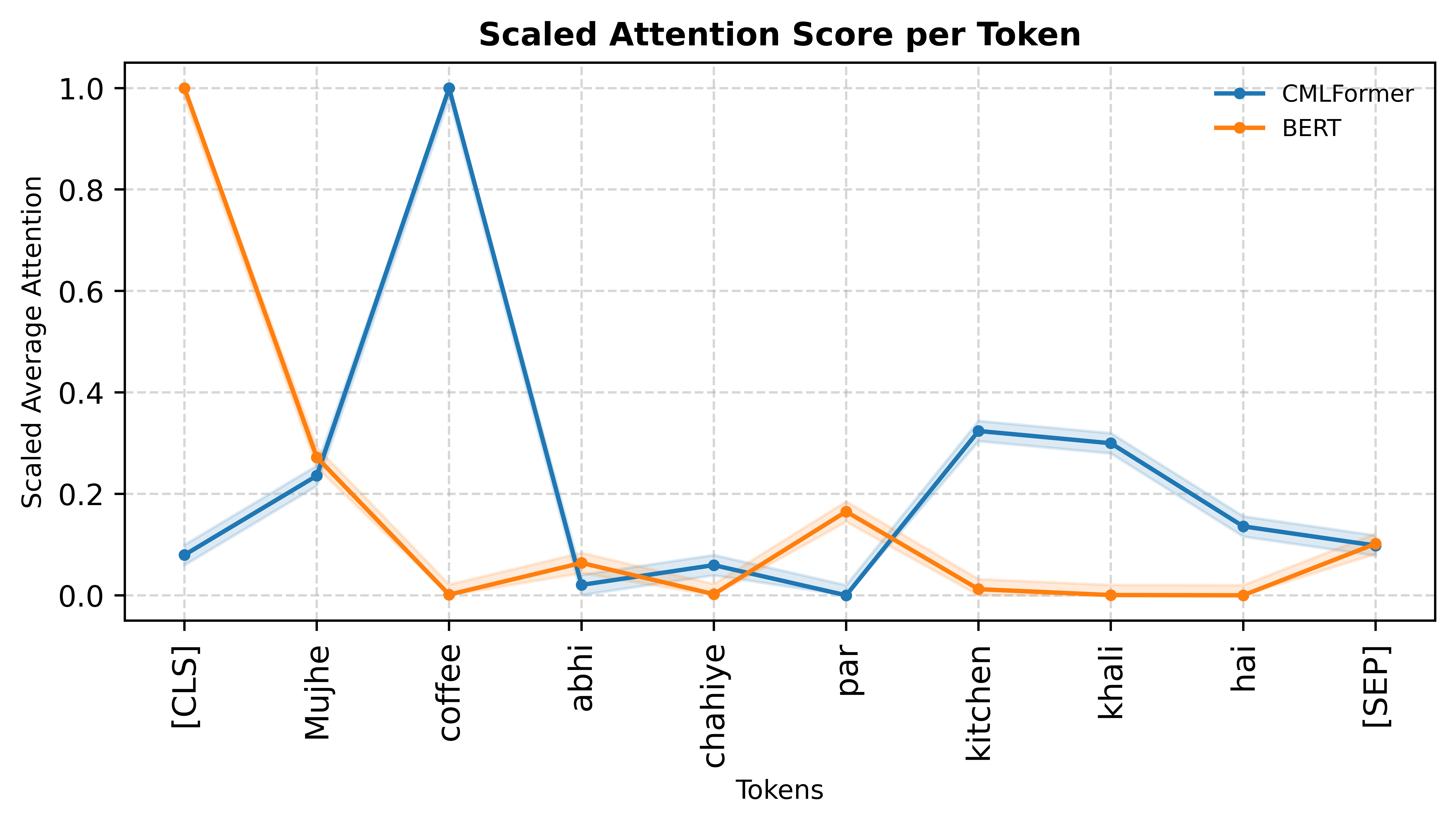}
\end{minipage}

\caption{\textbf{Average Attention Score per Token by CMLFormer and $\text{BERT}_{\text{base}}$}. We see that CMLFormer consistently identifies and attends to language transitions around switching points, and is agnostic to the nature and number of transitions; BERT fails to identify any transitions and attends to trivial tokens.}
\label{fig:att_main}
\end{figure*}

\section{Limitations}

Despite the architectural innovations introduced by CMLFormer, our current work is subject to several limitations. The primary constraint was the size of the pre-training dataset; due to computational resource and time restrictions, we were only able to pre-train on a small subset of 10,000 augmented samples from the L3Cube-HingCorpus. The limited data availability invites concerns of overfitted pre-training, potentially leading to a substantial loss of robust generalization abilities in CMLFormer. This likely led to an imbalance where the downstream evaluation dataset (HASOC 2021, with over 5,000 examples) could overpower the learned representations, potentially resulting in overfitting to the evaluation task, rather than specializing from pre-training.

Additionally, while our architecture employs a dual-decoder setup with a shared encoder, the encoder in this work was relatively underparameterized compared to the size of the dual decoders and the complexity of the pre-training tasks. This mismatch may have restricted the encoder's capacity to fully leverage the multi-task supervision and handle the representational burden required for effective cross-lingual alignment. We hypothesize that for effective multi-task learning involving a shared encoder and $n$ multiple decoders, the size of the encoder must be approximately $n$ times the size of each decoder.


\section{Future Work}

In our future work, we plan to pre-train CMLFormer on the full L3Cube-HingCorpus dataset with 52.9M sentences, along with higher-quality augmentations. Scaling up the dataset would provide more diverse switching patterns and lexical variations, improving the model's ability to generalize. We will also integrate Rotary Positional Encodings that can explicitly incorporate switching point information into token representations. This enhancement could allow the model to structurally internalize language transitions, providing stronger inductive bias toward code-mixed syntactic patterns. Future work will also explore scaling up the encoder's parameter count to match the dual decoders. Given that the encoder bears the majority of the representational and optimization load across multiple objectives, an asymmetric architecture with a larger encoder could significantly improve learning stability and downstream performance. 

\section{Conclusion}
In this work, we introduce \textbf{CMLFormer}, an enhanced dual–decoder Transformer that explicitly targets the linguistic structure of code–mixed text. We also introduce new pre-training objectives with a focused set of supervised and unsupervised tasks targeted at code-mixed language modeling - \textit{Masked Language Modeling} (MLM), \textit{Bilingual Language Translation Modeling} (BiLTM), \textit{Switching Point Prediction} (SPP), \textit{Bilingual Translated Sentence Prediction} (BTSP), \textit{Token Language Classification} (TLC) and \textit{Code–Mix Index Regression} (CMI). Through a combination of pre-training tasks, we aim to learn rich contextualized representations of code-mixing that are sensitive to both cross-lingual semantics and local language transitions.  

CMLFormer outperforms known baselines like \textsc{BERT}\textsubscript{base} on the HASOC-2021 Hinglish benchmark for hate-speech detection by yielding an absolute F1 gain of \textbf{0.18} and \textbf{0.05} through its BiLTM and SPP tasks, respectively. Attention-map analysis further shows that CMLFormer is far more capable than other approaches at effectively learning switching point dynamics in code-mixed inputs, indicated by higher attention scores obtained by attending higher to switching points caused by language transitions. Our extrinsic and intrinsic evaluations confirm our hypothesis and reinforce our claims that CMLFormer's architectural novelties and design make it powerful at code-mixed language modeling, enabling it to learn richer and more informative representations that internalize the syntactic and semantic nuances of code-mixed text and language transition cues.



\newpage

\bibliography{anthology,custom}
\bibliographystyle{acl_natbib}

\clearpage

\appendix

\section{Appendix}
\label{sec:appendix}

\subsection{Dataset Augmentation and Curation}
\label{app:dataset-aug-prep}

\subsubsection{Generating a Parallel Hinglish-English-Hindi Corpus}
To support the pre-training objectives, we proceeded with the data augmentation process with an initial sample of 10,000 Hinglish lines selected from the L3Cube-HingCorpus as a first milestone. The augmentation pipeline involved various stages applied to these 10,000 lines. First, the data was cleaned to remove emojis, special Unicode characters, and extra whitespace. Then, they were fed into Gemini 2.0 Flash to generate parallel translations in English and Hindi (Roman script).

\subsubsection{Augmentation Details and Example Transformations}

Below is an example from our augmented dataset showing Hinglish sentences alongside their English and Hindi (Roman) translations:

\medskip
\noindent\textbf{Example 1:}
\begin{itemize}
    \item \textbf{Hinglish:} \textit{aapki logo ki help krne ki soch bhut acchi h ...manav seva bhut hi accha kary krte ho di.}
    \item \textbf{English:} \textit{Your thought of helping people is very good... you do a very good deed of human service, sister.}
    \item \textbf{Hindi (Roman):} \textit{Aapki logo ki madad karne ki soch bahut achchhi hai... manav seva bahut hi achchha karya karte ho di.}
    \item \textbf{Labels:} \textit{[0, 0, 0, 1, 0, 0, 0, 0, 0, 0, 1, 1, 0, 0, 0, 0, 0, 0, 0]}
    \item \textbf{Switching Points:} \textit{[0, 0, 0, 1, 1, 0, 0, 0, 0, 0, 1, 0, 1, 0, 0, 0, 0, 0, 0]}
\end{itemize}

\medskip
\noindent\textbf{Example 2:}
\begin{itemize}
    \item \textbf{Hinglish:} \textit{Tum abhi Twitter use kr rhi ho :) }
    \item \textbf{English:} \textit{Are you currently using Twitter? :) }
    \item \textbf{Hindi (Roman):} \textit{Tum abhi Twitter use kar rahi ho? :) }
    \item \textbf{Labels:} \textit{[0, 0, 1, 1, 0, 0, 0, 0]}
    \item \textbf{Switching Points:} \textit{[0, 0, 1, 0, 1, 0, 0, 0]}
\end{itemize}

\subsubsection{Switching Point Labels Generation}  
To facilitate the Switching Point Prediction (SPP) objective, we generated fine-grained switching point labels for each tokenized Hinglish sentence in our augmented dataset. We approached this task by developing a semi-automatic labeling pipeline using Gemini 2.0 Flash, prompted to perform token-level language annotation with strict rules for tagging each token as either Hindi (hi) or English (en).

Each Hinglish sentence was first tokenized into whitespace-separated tokens. Then, Gemini was instructed to annotate every token individually without modifying their original structure, preserving punctuation and special symbols. Tokens were labeled as [en] for English or uncertain words and [hi] for Hindi, Urdu, or Punjabi words. Ambiguous cases defaulted to the previous token's language following a consistency-first principle.

After obtaining the annotated sentences, we parsed the labeled outputs to derive two sequences:
\begin{itemize} \item \textbf{Token Language Labels}: A sequence of 0s and 1s, where 1 denotes an English token and 0 denotes a Hindi token. \item \textbf{Switching Points Array}: A binary array computed by scanning the token labels and marking 1 whenever a language switch occurred between adjacent tokens, and 0 otherwise. \end{itemize}

For example, a labeled sequence like \texttt{Phone[en] ko[hi] charge[en] karo[hi]} results in a token language array \texttt{[1, 0, 1, 0]} and a switching points array \texttt{[0, 1, 1, 1]}. This allows CMLFormer to directly supervise on detecting fine-grained language transitions.

To ensure data quality, we applied post-processing filters to discard entries where the annotated labels were incomplete, ambiguous, or invalid (e.g., containing unexpected tags or lacking parallel translations). The final curated dataset contains the Hinglish sentence, its parallel English and Hindi translations (in Roman script), the token-level language labels, and the switching point labels, stored in JSONL format for pretraining. This process produced high-quality supervision data to explicitly teach the model to recognize and anticipate language switching points..

\subsubsection{Data Curation and Pre-processing}

During augmentation, we conducted a manual review of several hundred samples to ensure the quality of translation. However, we observed examples where producing a valid translation could not be generated because of unintelligible text or highly ambiguous phrasing. Such training examples were omitted from the dataset.

\subsubsection{LLM Prompts and Configuration}
We used the Gemini-2.0-Flash\footnote{https://ai.google.dev/gemini-api/docs/models\#gemini-2.0-flash} model from Google's Generative AI suite via the official Python SDK. The model was configured with output type set to \texttt{application/json} to ensure structured output for easier parsing and sentence extraction. To make sure the model adheres to our request, we prompted it with a few-shot examples.

The prompt we use is:
\begin{quote}
\ttfamily
\scriptsize
Task: Translate the given Hinglish text into both formal English and standardized Hindi (written in Roman script).

Input Hinglish: "\{hinglish\_text\}"

Requirements:

1. English translation should be grammatically correct and natural.\\
2. Hindi translation must use ONLY Roman script (Latin alphabet), not Devanagari.\\
3. Maintain the original meaning and tone in both translations.\\
4. Use standard transliteration conventions for Hindi.\\
5. Preserve context from the original text.

Important:

- DO NOT include any explanations, notes, or additional text.\\
- Respond ONLY with the exact JSON format shown below.\\
- Both translations should be complete sentences with proper punctuation.\\
- If the input text is not in Hinglish, English, or Hindi, return "NULL"\\
- If the input is Hindi, return the Hindi as is in the "hindi\_roman" field.\\
- If the input is English, return the English as is in the "english" field.

Return this exact JSON structure:

\{
  "english": "Your English translation here",\\
  "hindi\_roman": "Your Hindi translation in Roman script here"
\}

Examples:

Input: "Main kal movie dekhne jaa raha hoon"\\
Output: \{"english": "I am going to watch a movie tomorrow", "hindi\_roman": "Main kal film dekhne ja raha hoon"\}

Input: "Office ke baad hum coffee shop par milenge"\\
Output: \{"english": "We will meet at the coffee shop after office", "hindi\_roman": "Karyalay ke baad hum coffee shop par milenge"\}
\end{quote}

We chose a default temperature of 0.7 for generating translations to ensure translations are natural yet accurate. We also added a retry logic of 3 requests, with a delay of 2 seconds to avoid rate limiting. 

\subsection{Architecture Ablations}
\label{app:architecture-ablation}

\subsubsection{Dual Decoder without Decoder-level Cross Attention}
\label{app:dual-decoder-no-cross-attention-ablation}
To isolate and measure the impact of the proposed decoder cross-attention sub-layer in each layer, we will decouple the two decoders by removing cross-attention from both branches. This would degenerate both decoders to the vanilla architecture and remove the need for synchronous decoding since the overall architecture would function as two independent decoders with a shared encoder. While the model retains the multi-target training setup, this approach prevents each decoder from learning from the other, inhibiting cross-lingual alignment and learning inter-language relationships.

\subsubsection{Ablating with Cross-Attention Inputs}
\label{app:cross-attention-inputs-ablation}
We will investigate the effect of swapping the inputs to the cross-attention sub-layer with the final output hidden state from the previous layer of the other decoder after being passed through all its sub-layers. We hypothesize that this will ensure that each layer's cross-attention is based on a \textit{stable} and \textit{richer} representation from the other decoder branch. Denoting the output hidden states of the previous layer of the base and mixing language decoder as $H_{l-1}^b$ and $H_{l-1}^m$ respectively, this can be formulated as,

\begin{equation}
\begin{split}
o_{l}^b &= {Attention}(a_l^b, H_{l-1}^m, H_{l-1}^m) \\
o_{l}^m &= {Attention}(a_l^m, H_{l-1}^b, H_{l-1}^b)
\end{split}
\end{equation}

Output attention scores $o_{l}^b$ and $o_{l}^m$ are then passed as input to the feed-forward and layer normalization sub-layers as done previously. After each layer $l$ in the decoder, we store the final output hidden states $H_{l}^b$ and $H_{l}^m$ so they can be used in the cross-attention sub-layer of the decoder's $l+1$ layer.

\subsection{Synchronous Dual-Decoder Cross Attention Outperforms other Setups}

\begin{figure}[h]
\centering
\includegraphics[width=\linewidth]{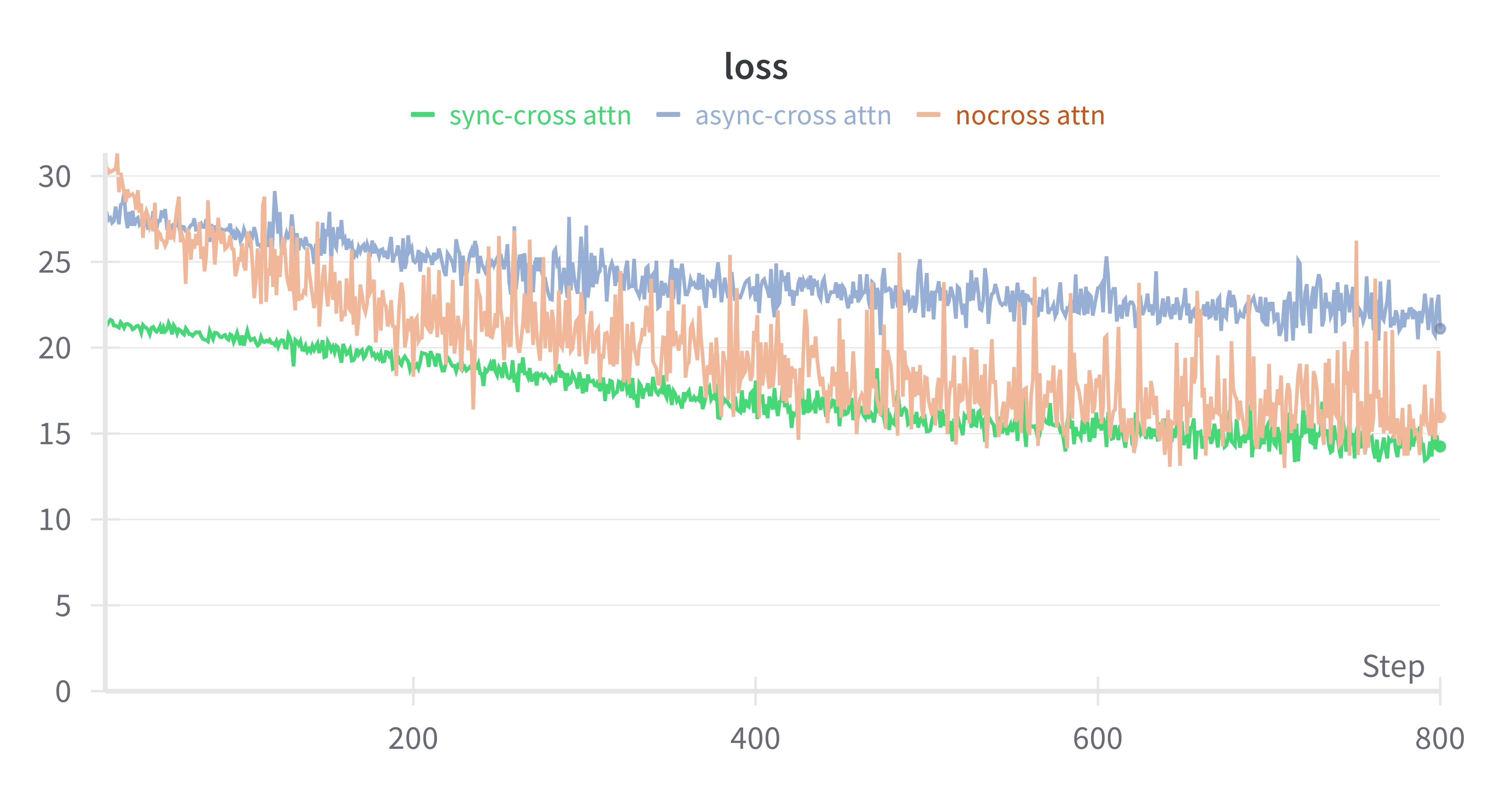}
\caption{\textbf{Loss curve for the total pre-training loss aggregated across all active objectives across different variants of decoder cross-attention}.  The lower curve for synchronous cross-attention reflects the strength of interaction learned between base and mixing language representations.}
\label{fig:dcrossattn_loss}
\end{figure}

We explored the three variants of decoder-level cross-attention through different pre-training runs. The synchronous variant, where each decoder attends to the current layer's hidden state from the other decoder, shows the smoothest and lowest loss trajectory consistently across our preliminary experiments. This exhibits effective mutual learning between the base and mixing language decoders and supports our hypothesis that fully synchronized decoders facilitate stronger cross-lingual alignment. 

The no cross-attention setup initially decreases in loss but exhibits high variance and noisy training throughout. Despite almost matching the synchronous cross-attention variant and consistently outperforming the asynchronous cross-attention setup, the instability implies poor generalization and inconsistent learning.

In contrast to our hypothesis, asynchronous cross-attention, which uses the previous layer's representations, leads to higher and much slower-converging loss, possibly due to delayed exchange of information leading to weaker inter-decoder interaction and suboptimal alignment.


\subsection{Pre-training Details}
\label{app:pretraining-details}

\subsubsection{Tokenizer}
The tokenizer was trained using standard WordPiece preprocessing and vocabulary construction techniques, with a vocabulary size of 32,000, following conventions established in models like BERT. This size offers a good trade-off between vocabulary coverage and computational efficiency and is especially well-suited for code-mixed data where subword-level generalization is important. 

\subsubsection{Model Configuration}

The model configuration for pre-training $\text{CMLFormer}_{\text{base}}$, which matches the number of parameters in $\text{BERT}_{\text{base}}$ is given in Table \ref{tab:model-config-base}.

\begin{table}[h]
\centering
\begin{tabular}{l l}
\hline
\textbf{Metric} & \textbf{Value} \\
\hline
Source vocabulary size & 32,000 \\
Base target vocabulary size & 32,000 \\
Mixed target vocabulary size & 32,000 \\
Number of layers & 12 \\
Hidden dimension & 768 \\
Feed-forward network dimension & 3,072 \\
Number of attention heads & 12 \\
Dropout rate & 0.1 \\
Maximum sequence length & 512 \\
Decoder cross-attention & \checkmark \\
Decoder cross-attention type & synchronous \\
\hline
\end{tabular}
\caption{Model configuration for $\text{CMLFormer}_{\text{base}}$.}
\label{tab:model-config-base}
\end{table}

\subsubsection{Pre-training Hyperparameters}

The best pre-training hyperparameters, including the loss weights assigned to each pre-training objective, are provided in Table \ref{tab:pre-training-args-base}.

\begin{table}[h]
\centering
\begin{tabular}{l l}
\hline
\textbf{Parameter} & \textbf{Value} \\
\hline
Epochs & 20 \\
Learning Rate (initial) & 1e-5 \\
Learning Rate Scheduler & Exponential Decay \\
Learning Rate Decay Factor & 0.9 \\
$\mathcal{L}_{\mathrm{MLM}}$ weight ($\alpha$) & 1.0 \\
$\mathcal{L}_{\mathrm{SPP}}$ weight ($\beta$) & 1.0 \\
$\mathcal{L}_{\mathrm{BTSP}}$ weight ($\gamma$) & 10.0 \\
$\mathcal{L}_{\mathrm{BiLTM}}$ weight ($\eta$) & 1.0 \\
$\mathcal{L}_{\mathrm{TLC}}$ weight ($\zeta$) & 10.0 \\
$\mathcal{L}_{\mathrm{CMI}}$ weight ($\delta$) & 1.0 \\
\hline
\end{tabular}
\caption{Pre-training hyperparameters and objective loss weights for $\text{CMLFormer}_{\text{base}}$.}
\label{tab:pre-training-args-base}
\end{table}

\subsection{Pre-Training Tasks and Training Losses}

\subsection{Switching Point Prediction Implementation Details}
\label{app:spp-details}

The switching point labels $T$ are provided at the word level, while the model operates on subword token sequences produced by the tokenizer. To ensure correct supervision, we align the switching labels to the tokenized inputs by assigning each switching label only to the first subword token of its corresponding word. All other subword tokens within the same word, as well as special tokens (e.g., padding), are masked out using a label of $-100$ (ignored index in PyTorch).

The alignment is performed by tracking token-to-word mappings provided by the tokenizer and applying switching labels only when a new word boundary at the first token of a word is encountered. This prevents duplicated supervision across subwords and ensures that loss computation reflects the original word-level switching annotations accurately.

\subsection{Fine-tuning Hyperparameters}
\label{app:finetuning-details}

The best fine-tuning hyperparameters are provided below.

\begin{table}[h]
\centering
\begin{tabular}{l l}
\hline
\textbf{Parameter} & \textbf{Value} \\
\hline
Epochs & 30 \\
Learning Rate (initial) & 1e-5 \\
Learning Rate Scheduler & Exponential Decay \\
Learning Rate Decay Factor & 0.9 \\
\hline
\end{tabular}
\caption{Fine-tuning hyperparameters for $\text{CMLFormer}_{\text{base}}$.}
\label{tab:fine-tuning-args-base}
\end{table}  

\subsection{Additional Attention Visualizations}
\label{app:attention-appendix}
\begin{figure*}[b]
\centering

\begin{minipage}{0.47\textwidth}
    \centering
    \includegraphics[width=\linewidth]{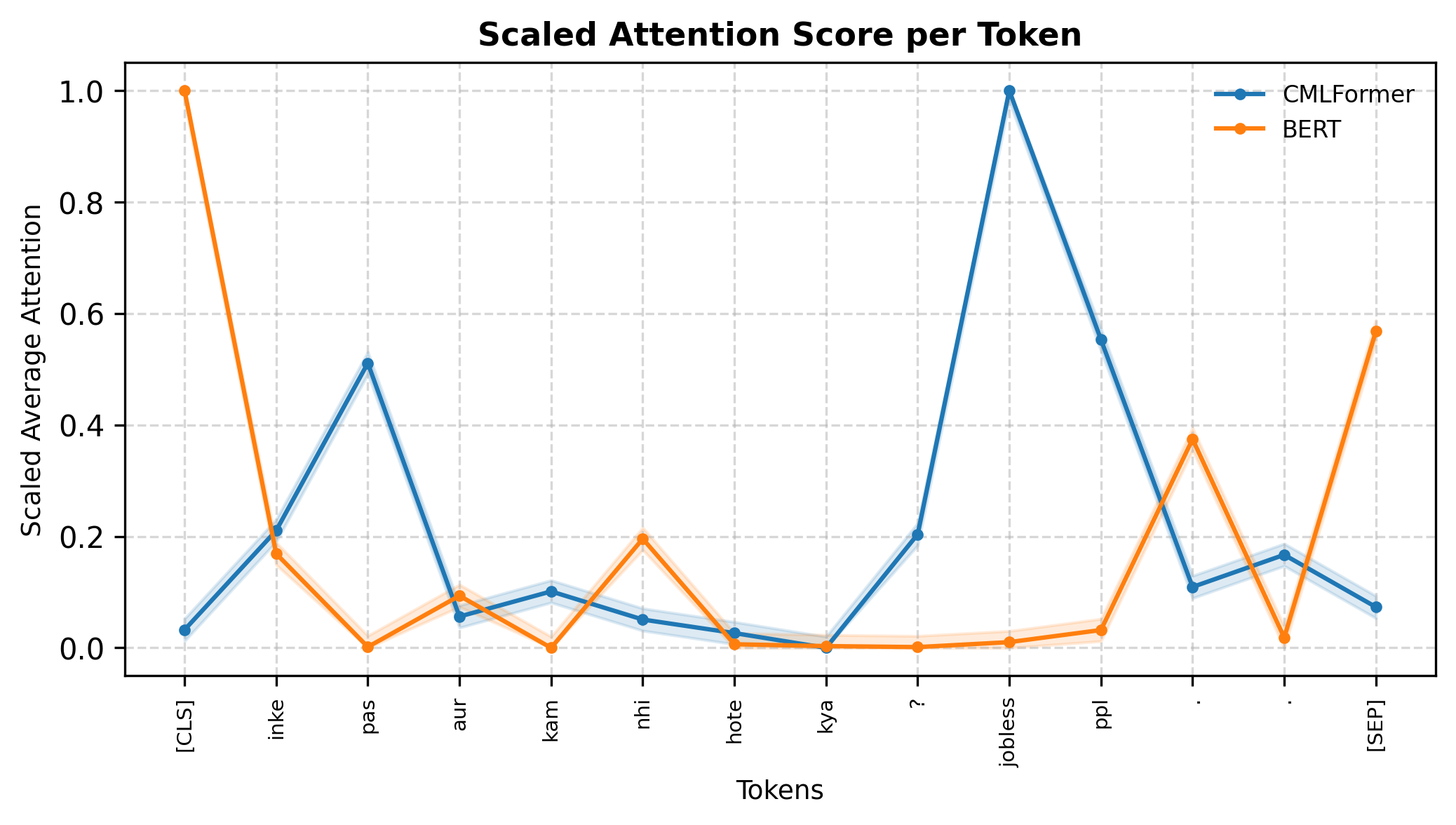}
\end{minipage}
\hfill
\begin{minipage}{0.47\textwidth}
    \centering
    \includegraphics[width=\linewidth]{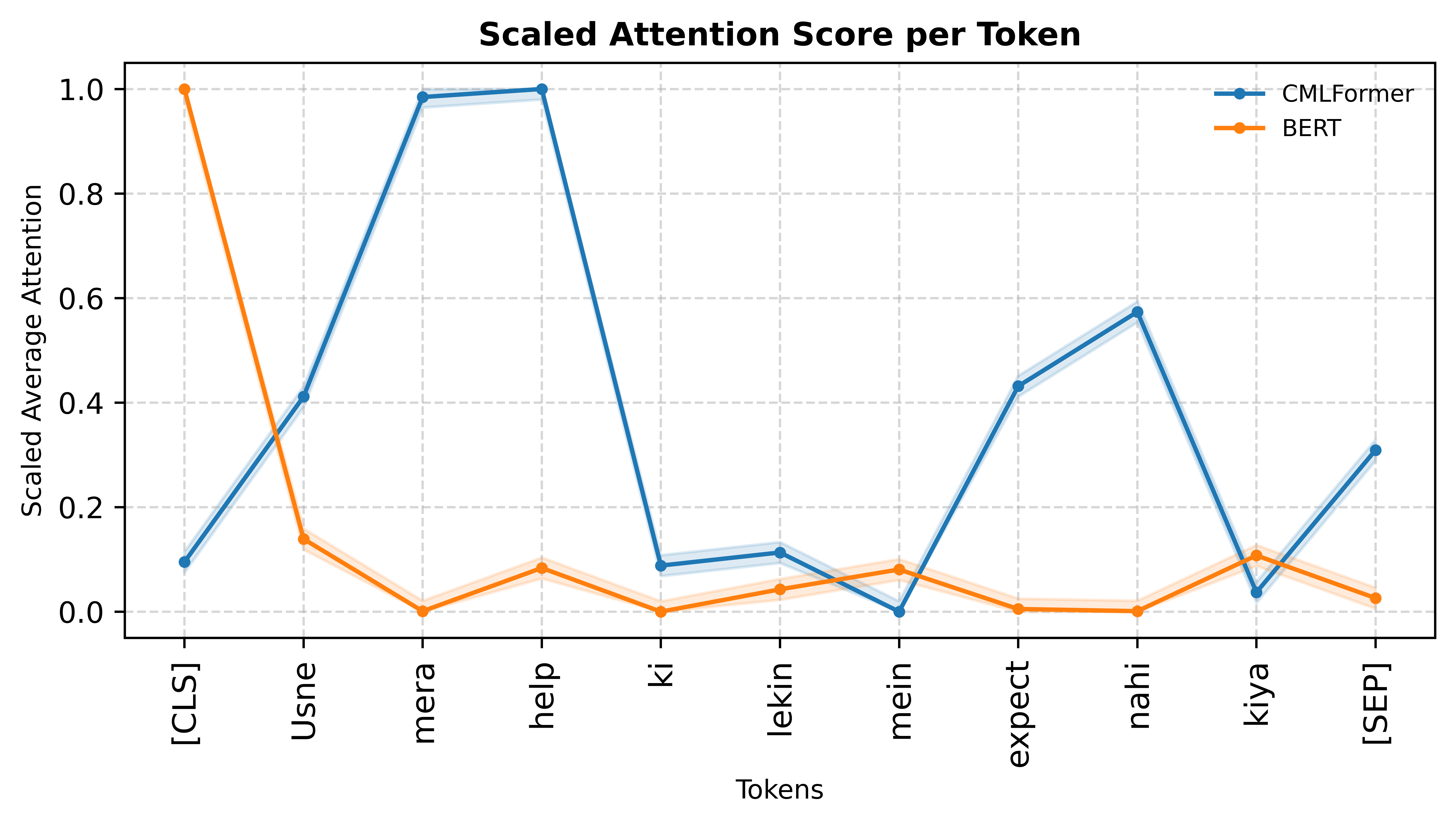}
\end{minipage}

\vspace{1em}

\begin{minipage}{0.47\textwidth}
    \centering
    \includegraphics[width=\linewidth]{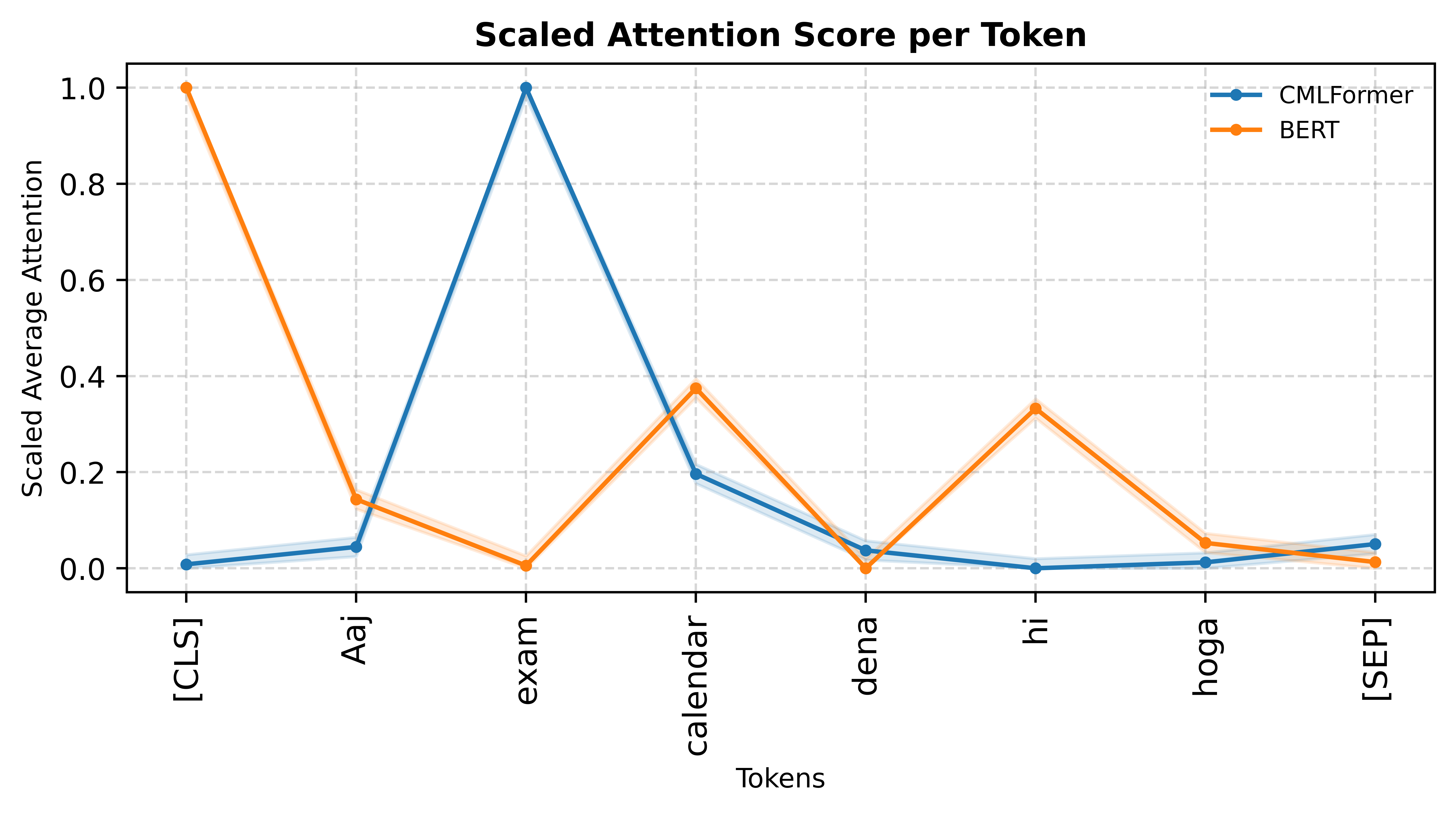}
\end{minipage}
\hfill
\begin{minipage}{0.47\textwidth}
    \centering
    \includegraphics[width=\linewidth]{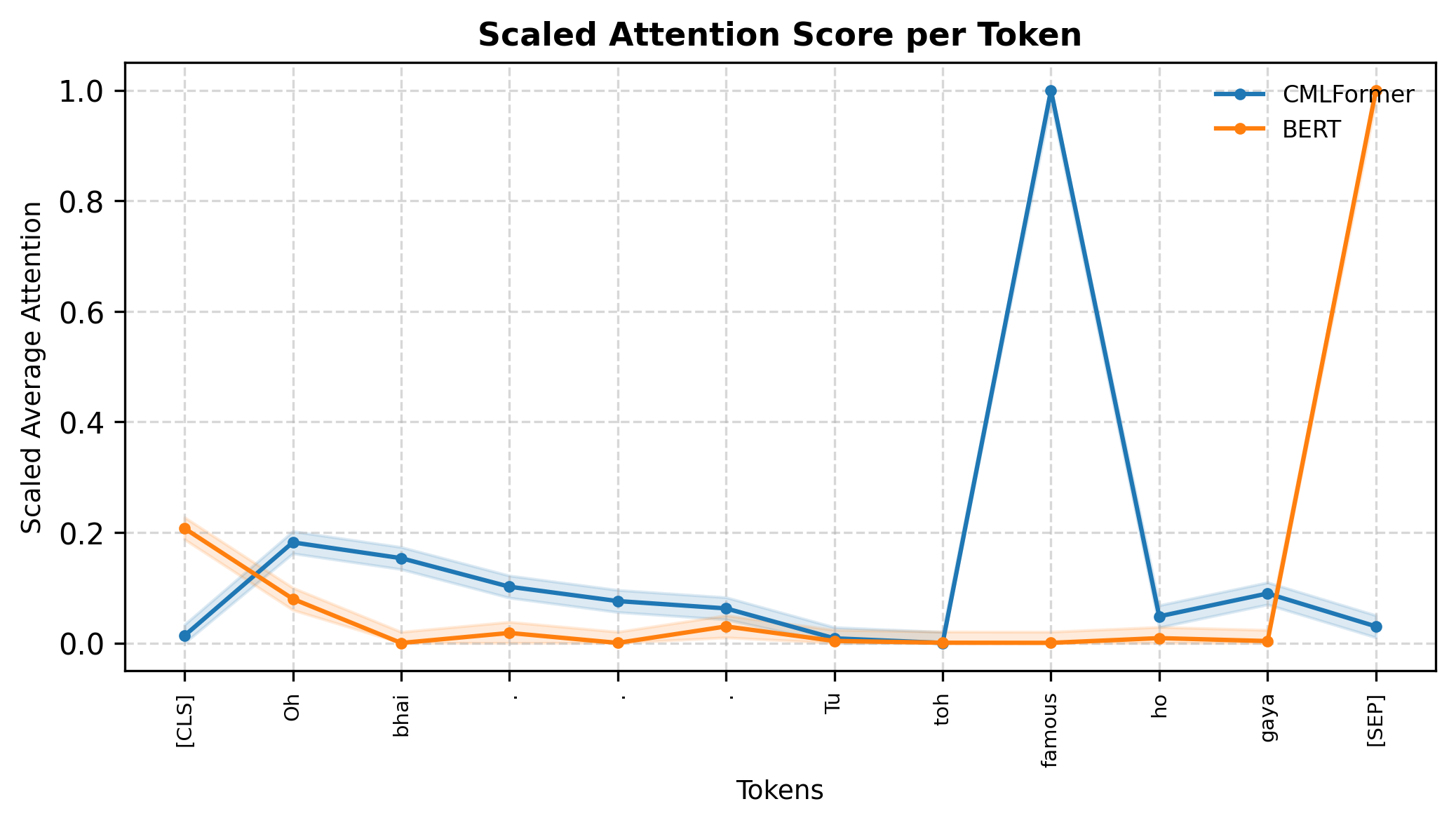}
\end{minipage}
\caption{\textbf{Additional attention maps produced by CMLFormer on code-mixed inputs.} Attention consistently peaks near switching points.}
\label{fig:att_appendix}
\end{figure*}

Additional attention maps generated by CMLFormer are shown in Figure \ref{fig:att_appendix}, complementing the examples provided in Section~\ref{sec:attention-analysis}. These plots further illustrate the model's sensitivity to language transitions within code-mixed text.

\end{document}